\documentclass[%
 reprint,
 amsmath,amssymb,
 aps,
]{revtex4-2}

\usepackage{graphicx}% Include figure files
\usepackage{dcolumn}% Align table columns on decimal point
\usepackage{bm}% bold math
\usepackage{charter}
\usepackage{upgreek}
\usepackage{enumitem}
\usepackage{hyperref}

\begin{document}

% \preprint{APS/123-QED}

\title{Disentangling intrinsic motion from neighbourhood effects in heterogeneous collective motion}% Force line breaks with \\
\thanks{Supplementary videos available at: \url{https://www.dannyraj.com/obsinf-supp-info}}%

\author{Arshed Nabeel}
 \affiliation{Center for Ecological Sciences, Indian Institute of Science, Bengaluru}
 \altaffiliation[Also at ]{Department of Chemical Engineering, Indian Institute of Science, Bengaluru}%Lines break automatically or can be forced with \\
\author{Danny Raj M.}%
 \email{dannym@iisc.ac.in}
\affiliation{Department of Chemical Engineering, Indian Institute of Science, Bengaluru
}%

\date{\today}% It is always \today, today,

\begin{abstract}
Most real world collectives, including active particles, living cells, and grains, are heterogeneous, where individuals with differing properties interact. The differences among individuals in their intrinsic properties have emergent effects at the group level. It is often of interest to infer how the intrinsic properties differ among the individuals, based on their observed movement patterns. However, the true individual properties may be masked by emergent effects in the collective.
We investigate the inference problem in the context of a bidisperse collective with two types of agents, where the goal is to observe the motion of the collective and classify the agents according to their types. Since collective effects such as jamming and clustering affect individual motion, an agent's own movement does not have sufficient information to perform the classification well: a simple observer algorithm, based only on individual velocities cannot accurately estimate the level of heterogeneity of the system, and often misclassifies agents.
We propose a novel approach to the classification problem, where collective effects on an agent's motion is explicitly accounted for. We use insights about the physics of collective motion to quantify the effect of the neighbourhood on an agent using a \emph{neighbourhood parameter}. Such an approach can distinguish between agents of two types, even when their observed motion is identical.
This approach estimates the level of heterogeneity much more accurately, and achieves significant improvements in classification.
Our results demonstrate that explicitly accounting for neighbourhood effects is often necessary to correctly infer intrinsic properties of individuals.
\end{abstract}

\maketitle

%\tableofcontents

\section{Introduction}

With the advent of sophisticated imaging techniques and machine learning algorithms, many experimental studies in the field of complex systems and complex flows turn to computer vision to investigate the underlying dynamics of the individuals (agents). Examples include the study of the motion of an intruder and the flow of grains in granular flows \cite{Candelier2010, Kolb2014, Harich2011, Kumar2014}, hydrodynamics of droplet interactions in a microchannel \cite{Thorsen2001, Beatus2012, Shen2014, Jose2011, Bremond2011}, cell migration studies\cite{Tremel2009, Ariano2005, McLennan2012, Sharma2015}, motion of synthetic active particles\cite{MasoumehMousavi2019, Si2020}, dynamics of traffic flows \cite{Kamijo1999, Moussaid2012, Nicolas2019} and collectives \cite{Murino2017a}, and flocking behaviour of social organisms\cite{Ballerini2008, Katz2011, Calovi2018, jhawar2020noise}. The agents that make up the collective are tracked and their velocities are computed from the measurements. The motion of these agents is a result of both self-propulsion (or external driving) and interactions between the agents. 
One of the primary goals in collective motion research is to understand the relationship between the observed dynamics, the intrinsic motion and the interaction effects of the agents in the collective. 

Most studies of collective phenomena assume that the individual agents in a collective are identical. However, in real-world systems, this is seldom the case. 
In material-collectives like droplets driven through a microchannel, differences in sizes could modify the level of confinement which alters the dynamics significantly\cite{Bremond2011}. 
It is not always possible to synthesise active particles with homogeneous morphology. For example, in the case of Janus particles, the inhomogeneity in the gold coating on these particles could produce gravitational torques that affect how they cluster\cite{MasoumehMousavi2019}.
In cellular systems, one finds heterogeneity to arise dynamically in the form of leader--follower states that dictate the migration dynamics of these cells, especially during processes like wound closure\cite{Qin2021}.
Moving beyond examples from soft matter; in animal groups, heterogeneity can arise due to the differences in age, sex or behavioural tendencies of the individuals \cite{Delgado2018Variation, Dyer2008ShoalComposition, Jolles2020Heterogeneity} all of which can reflect in the movement patterns of the individuals. 
% Animals can also form multi-species flocks where each species may have different movement properties \cite{Ward2018MixedSpecies, Sridhar2018Friendship}. 
In pedestrian dynamics, people in the crowd with different mobility, or with different destinations could significantly alter the dynamics of the crowd \cite{Zhang2016,Geoerg2021}.
In all these systems, it is usually of interest to identify the individual differences among agents based on their movement information and understand the impact on collective motion. 

However, the question of how well one can make such inferences, and how the collective dynamics affect the inferences, is relatively unexplored, as the collective dynamics of the system depend on the heterogeneity in interesting ways. Schumacher et. al. \cite{Schumacher2017} demonstrate that it is hard to quantify the actual heterogeneity of the agents in the system: they find that higher inter-agent interactions could exaggerate the true level of heterogeneity while confinement can do the opposite. 

The problem of inferring properties of a heterogeneous collective can be posed at three different levels:

\begin{enumerate}
    \item \emph{Estimating the level of heterogeneity:} At the coarsest level, one may need to estimate some measure of the level of heterogeneity in the collective and characterize the system on a scale ranging from fully homogeneous to highly heterogeneous.
    \item \emph{Characterizing individual agents:} A harder problem would be to characterize individual agents according to their properties. For example, it might be possible to classify individuals into different types according to some intrinsic characteristics.
    \item \emph{Recovering intrinsic properties and governing dynamics:} At the highest level, one would be wish to recover the full intrinsic properties the individual agents or reconstruct the governing equations of motion for the individuals. This is a significantly harder problem.
\end{enumerate}

To investigate these questions, we turn to a simple heterogeneous collective: a \emph{bidisperse collective}, which consists of two distinct groups of agents. The agent properties are set to be identical within a group but are distinct between groups. 
Although simple in principle, such bidisperse collectives come up in a variety of fields, examples including migrating cells \cite{kwon2019stochastic, Blanchard2019Mesoscale}, oppositely driven systems like charged colloids or dusty plasmas \cite{Vissers2011, Leunissen2005, Sutterlin2009}
pedestrian crowds \cite{Helbing2005, Feliciani2016, Zhang2019Pedestrian} and mixed-species animal groups \cite{Sridhar2018Friendship, Ward2018MixedSpecies}.

We approach the problem of inferring a heterogeneous collective, by classifying agents into their groups in a bidisperse collective and estimating the group heterogeneity from the identified group memberships.
Depending on the context, fast and accurate classification can be very crucial. For example, while working with densely packed agents, it may be very hard to retain the tracking for long time windows and in problems like predicting the onset of a stampede, timely classification is critical in employing prevention strategies. 
The classification problem becomes particularly challenging when clean, labelled data is not readily available---as is often the case in real-world scenarios. In these situations, we cannot rely on the traditional supervised learning techniques \cite{hastie2009overview} for classification. A physics-informed approach, based on phenomenological understanding of the collective dynamics, is appealing in this context. By incorporating insights about the dynamics, such an approach can reduce the dependence on labelled data.

We begin with a description of a model for a bidisperse collective and show how, despite being simple, it gives rise to a wide range of collective phenomena which are interesting from the context of inferring the heterogeneity of the collective.
We first examine the performance of a simple observer that takes into account only the dynamics of a focal agent; because it is reasonable to expect that the information that aids the classification of a particular agent is encoded in the motion of that agent itself.
However, since the dynamics of an agent is also driven by collective effects arising from interactions with other agents, this approach often produces misclassifications. We address this problem by explicitly accounting for the collective effects on the movement of an individual. Based on the physics of collective motion, we derive a \emph{neighbourhood parameter} that quantifies the influence of neighbours on the movement of the focal agent; and classify agents based on how the agent's velocity compares to the neighbourhood parameter. This enables us to distinguish between agents of different groups, even when their observed movement is identical. Our analysis presents a physics-inspired approach to the inference problem, instead of a supervised-learning approach which is reliant on the availability of labelled training-data.

\section{Model and Dynamics} \label{sec:dynamics}

\subsection{A simple model of a bidisperse collective}

\begin{figure}
    \centering
    \includegraphics[width=\linewidth]{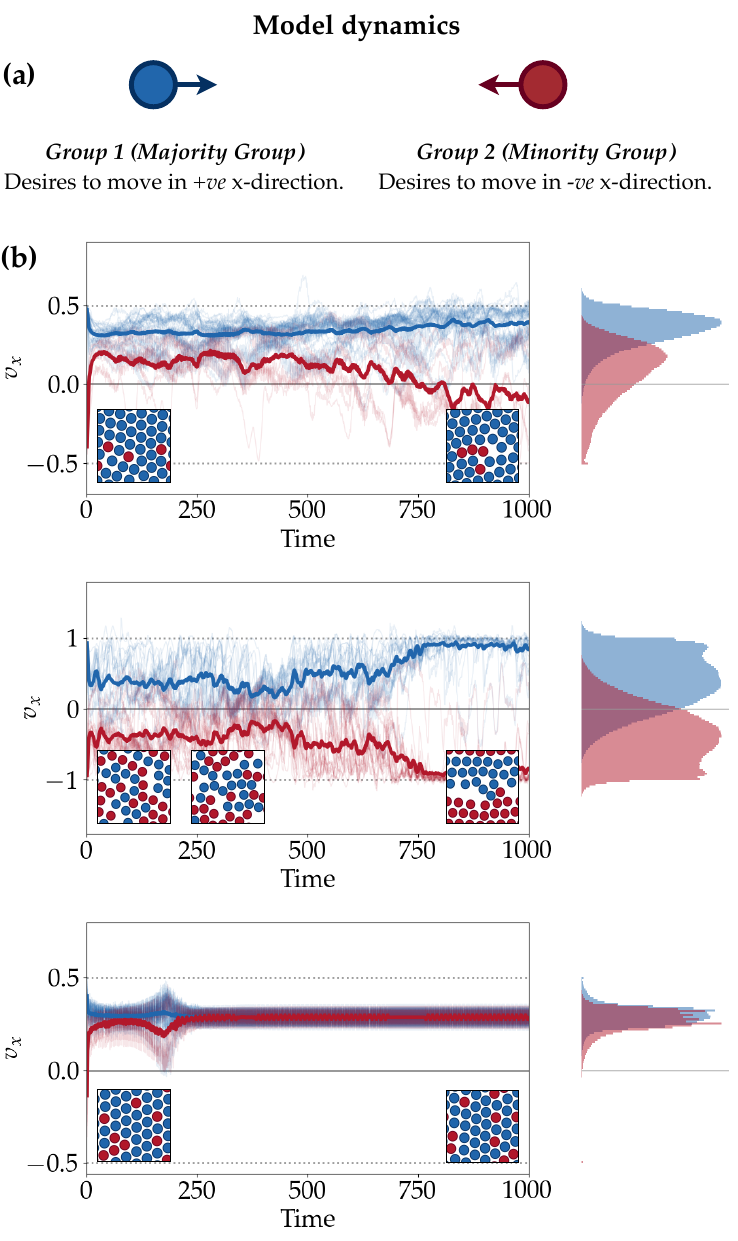}
    \caption{The bi-disperse collective system exhibits a wide range of dynamics.
    (a) The system consists of two different groups of agents, each with its own desired direction of motion.
    (b) Examples of dynamics exhibited by the model at different parameter values. Plots show the horizontal velocities ($\mathbf{v}_{i, j} \cdot \mathbf{e}_x$) for the two groups of agents from a single realization. The faint blue and red traces are the individual agent velocities for Group 1 and Group 2 agents respectively. The thick blue and red traces are the average velocities (averaged over all agents in the group) for Group 1 and Group 2 agents. The dotted grey lines show $\pm s_0$, the desired $x$-velocities for the two groups. Insets show snapshots of the model dynamics at different time-points.
    The histograms show the distribution of agent velocities for the two groups, across all time-points, and across 100 realizations. Notice that the histograms have significant overlap, and histograms of the minority group (red in the top and bottom plots) are significantly shifted from their respective $s_0$. (Also see Movie~S1) \emph{Top: } Example of clustering ($N_r = 3/14, \rho = 0.46, s_0 = 0.5$. \emph{Middle: } Example of lane formation ($N_r = 1/2, \rho = 0.46, s_0 = 1$). \emph{Bottom:} Example of jamming ($N_r = 3/14, \rho = 0.58, s_0 = 0.5$).
    }
    \label{fig:dynamics}
\end{figure}

We use a simple model of bidisperse collectives (i.e. consisting of two distinct groups of agents), similar to the models previously used in literature \cite{Helbing2000a, Reichhardt2018}. Circular agents with radius $R$, are arranged in a 2D periodic domain, with a packing density $\rho$. Each agent has a desired direction of movement, which is same for all agents in the group but different between the groups (which we call Group 1 and Group 2). Group 1 agents have a desired velocity along the positive $x$-direction and Group 2 agents in the negative $x$-direction. An agent is driven by two forces: a \emph{restitution force}, denoting the intrinsic effort by the agent to move in the desired direction, and an \emph{inter-agent force}, denoting the interactions between agents. The forces are designed such that the agents cannot overlap.

The following equations govern the dynamics of the agents:

\begin{eqnarray}
	\label{eqn:GovEq}
	m\frac{d\mathbf{v}_i}{dt}=\frac{m}{\tau} (\mathbf{v}_{0, i} - \mathbf{v}_i) + \sum_{\forall j \neq i} \mathbf{F}_{ij} \\
	\nonumber \\ 
	\label{eqn:Desdir}
	\mathbf{v}_{0, i}=\left\{\begin{matrix}
		+ s_0 \mathbf{e}_x & i \in \text{\textit{Group 1}}\\ 
		- s_0 \mathbf{e}_{x} & i \in \text{\textit{Group 2}}
	\end{matrix}\right. \\
	\nonumber \\ 
	\label{eqn:interagforce}
	\mathbf{F}_{ij}=\left\{ \begin{matrix} 
		-\gamma (d_{ij}-2R)^{-3} \hat{\mathbf{d}}_{ij} & d_{ij}<l_{cr} \\
		0 & \text{\textit{otherwise}}
	\end{matrix} \right. 
\end{eqnarray}

The intrinsic velocity $\mathbf{v}_{0, i}$ has a magnitude $s_0$, and is directed along the positive or negative x-direction depending on the group of the individual (see Equation~\ref{eqn:Desdir}). This is the source of heterogeneity in the model. 
The inter-individual force $\mathbf{F}_{ij}$ is a repulsive force, and decays as a power-law with distance. To prevent collisions, $\mathbf{F}_{ij}$ is chosen such that its magnitude blows up to infinity when the agent boundaries touch ($d_{ij} = 2R$). To avoid spurious interactions between far-off agents, $\mathbf{F}_{ij}$ is set to 0 beyond a cut-off radius, $l_{cr} = 3R$ (see Equation~\ref{eqn:interagforce}). $m$ is the mass of an individual agent (set to $1$), $\tau$ is the inertial time-scale of the system (set to $0.2$), and $\gamma$ determines the strength of the inter-agent interactions (set to $0.2$). $\mathbf{e}_x$ denotes the unit vector along x-direction. We define the packing density $\rho$ as the number of agents per unit area.

Key parameters that determine the collective dynamics include the packing density $\rho$ which we vary between $[0.22, 0.58]$, the intrinsic speed $s_0$ between $[0.1, 3]$ and the number ratio, the fraction of agents in the minority, $N_r$ between $[1/42, 1/2]$, which quantifies the degree of heterogeneity in the collective. We perform simulations for each combination of parameters $\{\rho, s_0, N_r\}$. Since, phenomena observed in bidisperse collectives such as laning, clustering, jamming, etc. depend upon the initial conditions of the simulations, we perform $100$ simulations for each set of parameter values where agent positions and their group-identities are assigned randomly.

\subsection{Model dynamics}

In the absence of any obstacles or collisions, an agent approaches its steady-state velocity $\mathbf{v}_{0,i}$ with a timescale $\tau$. However, when there are other agents present, the inter-agent interactions affect the movement in interesting ways. An agent can be blocked or pushed around by other agents in its path, which can result in diverse dynamics depending on the model parameters.
	
When the packing density $\rho$ is not too high (agents can freely move past each other), lateral migration due to interactions with other agents causes agents of the same group to find each other passively and form clusters (Figure~\ref{fig:dynamics}(b), top panel). The cluster of agents moves together as a unit, and collisions with the opposite group appear only at its boundary. For these reasons, clustered state is an absorbing state: i.e., once a cluster is formed, it does not break easily, but new agents can join the cluster. Clustering improves mobility of the agents, as the cluster as a whole is able to better force its way through opposing agents. Here, the word mobility is used to qualitatively describe the net movement of the agents in the collective: higher mobility implies larger movement of agents in their respective desired directions.
	
When $N_r$ is close to $\frac{1}{2}$ (symmetric or nearly symmetric regime), clustering can eventually lead to formation of system-spanning lanes (Figure~\ref{fig:dynamics}(b), middle panel). Since each lane consists only of one group of agents, mobility is maximum (agent speeds are close to $\pm s_0$) in the laned state.
	
When $\rho$ is high and the intrinsic velocity $s_0$ is relatively small, the collective can enter a jammed state, where agents meet head-on and do not have enough space to move past each other (Figure~\ref{fig:dynamics}(b), bottom panel). For the symmetric case ($N_r = \frac{1}{2}$), this causes the entire assembly to freeze. For asymmetric cases ($N_r < \frac{1}{2}$), due to an imbalance in the net force, the jammed assembly of agents slowly drifts in the direction dictated by the majority group. (See Movie~S1 for an animated visualization of these cases.)

Shown alongside the velocity plots are the distributions of the individual velocities for the two groups. For asymmetric cases, the minority group agents are often pushed in the opposite direction by the majority group. As a result, the velocity distribution for the minority group is shifted in the positive direction (see top and bottom panels). Besides, due to collective effects such as local jamming, there is a significant overlap between the velocity distributions of the two groups: for example, when mobility is very low, the two distributions are nearly indistinguishable (bottom panel). In other words, there can be agents of Group~1 and Group~2 moving with identical velocities.

\section{Observer Design and Classification}
In the context of our model collective, the three levels of the inference problem become the following:
\begin{itemize}
    \item \emph{Group heterogeneity level:} The number ratio $N_r$ is a direct measure of the level of heterogeneity in the group: heterogeneity is maximum at $N_r = 1/2$, and low when $N_r$ is low. Hence, at this level, our goal would be to estimate $N_r$.
        \item \emph{Classification level:} To classify agents into Group 1 or Group 2, we only need to ascertain the direction of $\mathbf v_{0, i}$, the magnitude is irrelevant.
    \item \emph{Discovering intrinsic drives:} This involves precisely identifying both the magnitude and direction of $\mathbf v_{0, i}$.
\end{itemize}

\noindent In this paper, we mainly focus on the first and second inference problems, and provide a broad discussion on how one could tackle the third, harder problem. 

An observer collects movement information about the agents, i.e. their positions and velocities, but has no other information about the details of the model. The observer then attempts to classify agents as Group~1 or Group~2 (Figure~\ref{fig:simple-obs}(a)). 

\subsection{Observing agents with a simple observer model}

A straightforward approach for classification is to classify agents based on their observed direction of motion. For a better estimate that eliminates transient fluctuations, one may use the average velocity computed over a time-window. Therefore, our simple observer algorithm proceeds as follows. First, the observer computes the average velocity $v_i^w$ of the agents, averaged over a time-window of length $w$.

\begin{equation}\label{eqn:observer_avg}
	v^w_i = \langle \mathbf{v}_i.\mathbf{e}_x \rangle_{w} 
\end{equation}

The observer classifies an agent based on the net direction of motion during this window, i.e. the sign of $v_i^w$:
\begin{eqnarray}\label{eqn:classifer1}
	\begin{matrix}
    	v^w_i \geq 0    &: \; \text{\textit{Group 1}} \\
    	v^w_i < 0       &: \; \text{\textit{Group 2}}
	\end{matrix}
\end{eqnarray}

\begin{figure*}
    \centering
    \includegraphics[width=\textwidth]{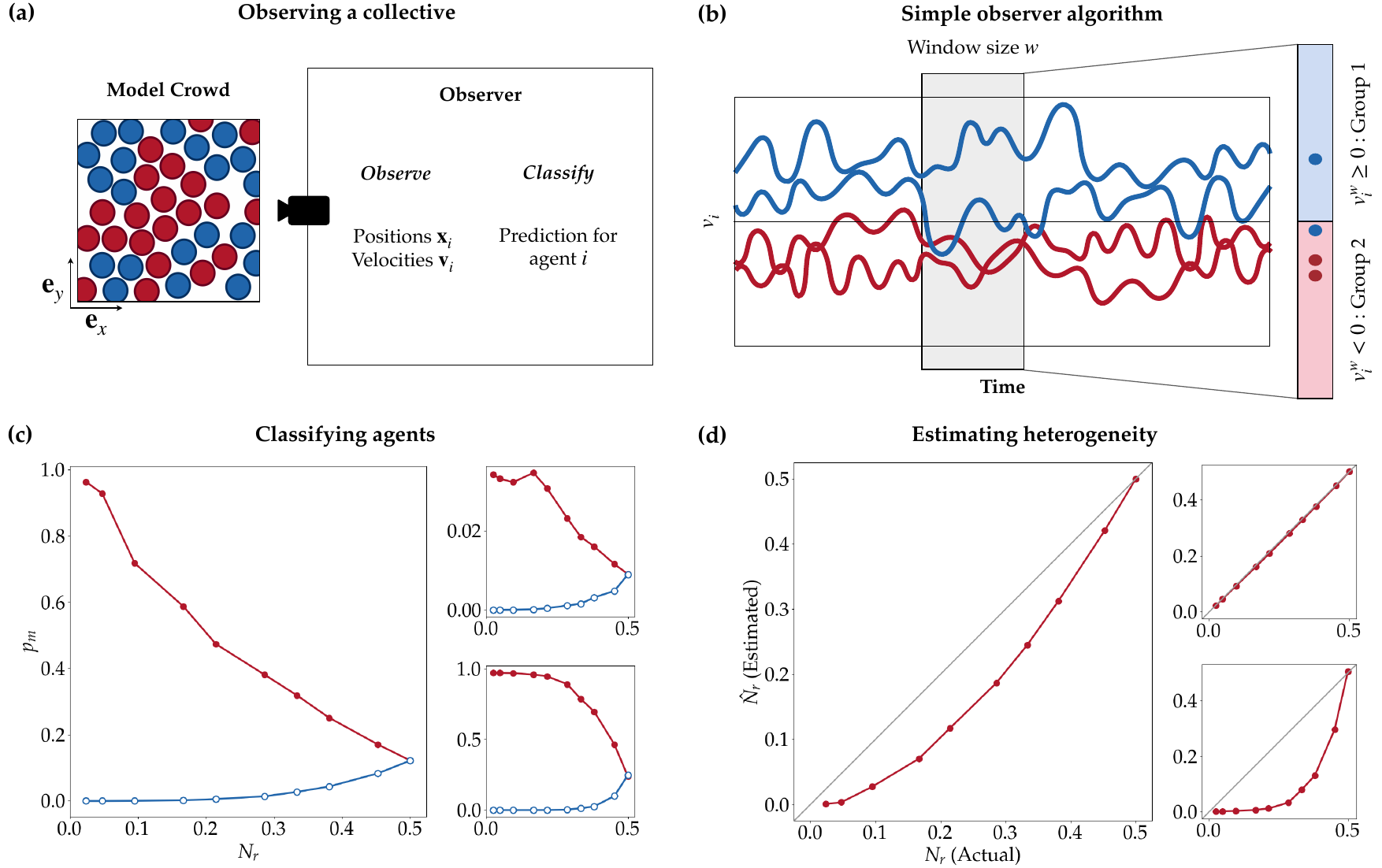}
    \caption{
    Classifying agents with a simple observer: the simple observer algorithm systematically misclassifies minority group agents and under-estimates the level of heterogeneity.
    (a) Illustration of the observer framework: the observer observes the positions and velocities of agents in the collective, and employs a classifier to classify agents as \emph{Group~1/Group~2}.
    (b) Classification algorithm for the simple observer: For each agent, the observer finds the average $x$-velocity of the agent within a short time window, and classifies the agent based on whether this velocity is positive or negative.
    (The box of the right shows a snapshot from the simulation, showing the Group~2 focal agent being pushed by Group~1 agents.)
    (c) The probability of misclassifying agents $p_m$ for the minority (red, closed circles) and majority (blue, open circles) groups, as a function of the number ratio $N_r$, for different levels of agent mobility (main: intermediate mobility, $\rho = 0.46, s_0 = 1$, top right: high mobility, $\rho = 0.31, s_0 = 2$, bottom right: low mobility, $\rho = 0.58, s_0 = 0.75$).
    $p_m$ is higher for lower $N_r$, denoting poorer classification performance for smaller groups.
    (d) The estimated value of $N_r$, which is a measure of estimated heterogeneity, as a function of true $N_r$, for different levels of agent mobility. Except when mobility is high, simple observer systematically underestimates $\hat N_r$.
    }
    \label{fig:simple-obs}
\end{figure*}

Figure~\ref{fig:simple-obs}(b) is a visual representation of the algorithm.

For a given window, the number of misclassifications is identified by comparing the labels of the agents predicted by the observer to the ground-truth labels (the actual desired directions for the agents, which are unknown to the observer). For each set of values for the parameters $N_r$ and $s_0$ we compute the probability of misclassifications ($p_m$), computed over multiple time windows and multiple independent realizations of the simulation. To estimate the level of heterogeneity in the system, we use the estimated value of $N_r$, i.e. the fraction of agents classified as Group 1 by the observer. 

This simple observer algorithm is the most straightforward way to classify agents. It assumes that the information necessary to classify an agent is fully contained in its own motion; and does not explicitly account for its interactions with the surroundings. The simple observer will serve as a baseline to study the effect of emergent collective dynamics on classification, before we develop improved classification models.

In a real-world scenario, an observer has to tackle the problem of sensor noise that may corrupt the observations. However, we make the simplifying assumption that the observer can observe the system perfectly without noise; the only unknown is the individual group identities of the agents, which the observer desires to infer. 

In our analyses, we used a window-length $w=50$. A shorter window corresponds to decision-making with fewer data points, while a larger window size may result in improved classification performance. The exact window length does not qualitatively affect the results, we verified this by repeating our analysis with several window sizes between $25$ and $100$.

\paragraph*{The simple observer mis-identifies minority group agents.}

The simple, velocity-based classification approach is most effective when the mobility is high ($s_0$ is high and $\rho$ is low) and agents are free to move in their own desired directions (Figure~\ref{fig:simple-obs}(c), top right panel). In general, however, agents will be pushed around by their neighbours and may not move in their desired directions at all times, which will lead to misclassifications. This effect is worse for the minority group---hence, the misclassification probability for the minority group is higher for smaller group sizes (Figure~\ref{fig:simple-obs}(c), main panel). This effect is even more pronounced in the case of very low mobility (low $s_0$ and high $\rho$) and low $N_r$, when almost all minority agents will be pushed in the wrong direction. Until $N_r$ reaches a critical threshold so that the agents can start to form clusters, all minority agents will be misclassified.

This effect is also visible in the estimated heterogeneity, $\hat N_r$. In general, the simple observer consistently underestimates the proportion of minority group agents except when the mobility is very high. When mobility is low, the observer estimates zero heterogeneity for small values of $N_r$; i.e. the observer cannot even detect the presence of minority agents.

\subsection{\label{sec:neigh-obs} Correcting misclassifications with a neighbourhood-based observer}

\begin{figure*}
    \centering
    \includegraphics[width=\textwidth]{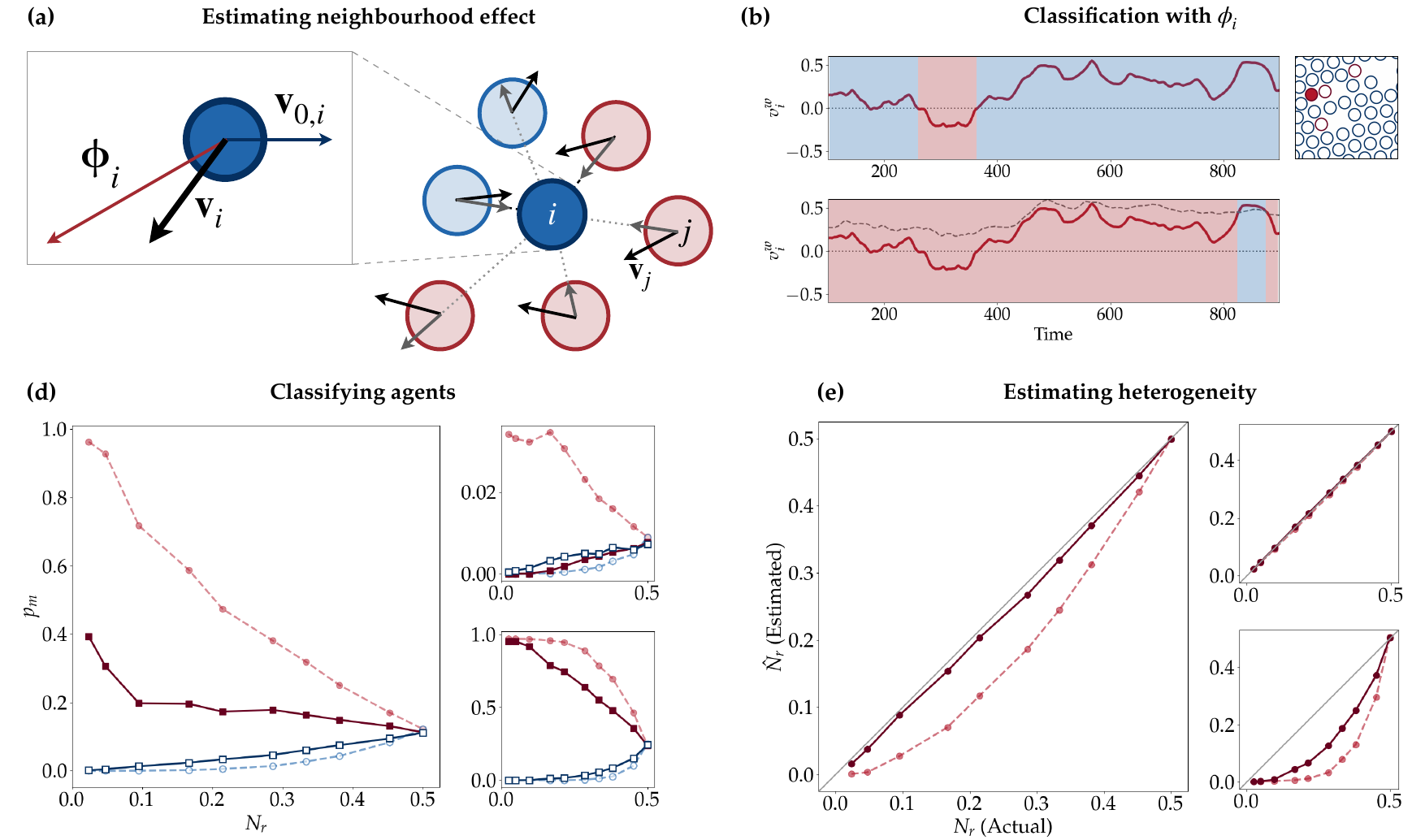}
    \caption{
    Classifying agents with a neighbourhood observer: an observer algorithm that accounts for neighbourhood effects significantly improves classification performance, and estimates heterogeneity more accurately.
    (a, \emph{box}) The velocity $\mathbf v_i$ of an agent can be decomposed into two components: $\mathbf v_{0, i}$, the intrinsic velocity of the agent, and $\boldsymbol{\upphi}_i$, the influence of neighbours. As seen, $\mathbf{v}_{0, i}$ and $\mathbf{v}_i$ may have different directions.
    (a) The neighbourhood parameter $\widehat{\boldsymbol{\upphi}}_i$ is an estimate of $\boldsymbol{\upphi}_i$, using velocities of neighbours. $\widehat{\boldsymbol{\upphi}}_i$ for an agent $i$ (deep blue) is determined by the velocities of its neighbours (pale blue and red). The contribution of neighbour $j$ is $\tilde{\mathbf{v}}_j$, the component of the velocity vector $\mathbf{v}_j$ towards $i$. $\widehat{\boldsymbol{\upphi}}_i$ is the mean of $\tilde{\mathbf{v}}_j$'s, scaled by an appropriate scaling factor.
    (b) Visualising the working of the neighbourhood classifier: 
    In the top panel, the red solid line is the observed velocity $v_i^w$ of a Group~2 agent. The shading corresponds to the predictions of the simple observer (\emph{Blue:} Group~1, \emph{Red:} Group~2). The simple observer misclassifies the agent as Group~1 whenever $v_i^w > 0$, which is often the case.
    In the bottom panel, the red solid line is the observed velocity $v_i^w$ and the dotted line is the neighbourhood parameter $\phi_i^w$. The shading corresponds to the predictions of the neighbourhood observer. The neighbourhood observer misclassifies the agent as Group~1 only when $v_i^w > \phi_i^w$, which happens only rarely.
    (c) The probability of misclassifying agents $p_m$ for the minority (red, filled squares) and majority (blue, open squares) groups, as a function of the number ratio $N_r$, for different levels of agent mobility (main: intermediate mobility, $\rho = 0.46, s_0 = 1$, top right: high mobility, $\rho = 0.31, s_0 = 2$, bottom right: low mobility, $\rho = 0.58, s_0 = 0.75$).
    The performance curves of the simple observer are shown as dotted lines. The neighbourhood observer drastically reduces misclassifications on the minority group.
    (d) The estimated value of $N_r$, which is a measure of estimated heterogeneity, as a function of true $N_r$, for different levels of agent mobility. The performance curves of the simple observer are shown as dotted lines. The neighbourhood observer estimates heterogeneity better than the simple observer, and can detect the presence of heterogeneity at a lower $N_r$.
    }
    \label{fig:neigh-obs}
\end{figure*}

The simple observer algorithm classifies agents based on their net velocity measured over a time window $w$. 
For instance, when the observed velocity $v_x^w$ of a Group~2 agent is positive, it will be classified as Group~1 incorrectly by the observer. Thus, correcting a misclassification would require the observer to classify the agent as belonging to Group 1 even if it exhibits a net movement in the negative direction.
From the perspective of an observer, this is a counter-intuitive step: the observer has to differentiate agents in Group 2 from those in Group 1 that happen to move in the same x-direction. We need to ascertain the conditions under which an observer should swap the identified group-identity of the agent to correct a possible misclassification event.

\paragraph*{An agent's neighbourhood encodes information about its intrinsic direction.}
In the absence of interactions, an agent moves in its desired direction of motion with velocity $s_0$, and classification is a trivial task. However, an agent's movement is influenced by both its own intrinsic motion, as well as interaction-effects from its neighbours which can aid or hinder the motion. For improved classification performance, an observer needs to decouple the component of an agent's motion due to its intrinsic drive and the component of motion caused by these interaction effects.

In the current formulation of the classifier, the observer uses $v_i$ as a proxy for the agent's desired velocity, $\mathbf{v}_{0,i}$. However, in reality, the observed $\mathbf{v}_{i}$ is a combination of the agent's effort to move in $\mathbf{v}_{0,i}$ and the support or resistance offered by the neighbourhood $ \boldsymbol{\upphi}_i $.

\begin{equation}\label{eqn:phen-neighfield}
	v_i = \mathbf{v}_{0,i} \cdot \mathbf{e}_x + \boldsymbol{\upphi}_i \cdot \mathbf{e}_x
\end{equation}

See Figure~\ref{fig:neigh-obs}(a) for an illustration. With this definition (Equation~\ref{eqn:phen-neighfield}), the new classification criterion becomes,
\begin{eqnarray}
    \begin{matrix}
    	\mathbf{v}_{0,i} \cdot \mathbf{e}_x = v_i - \phi_i \geq 0 & : \; \text{Group 1} \\
    	\mathbf{v}_{0,i} \cdot \mathbf{e}_x = v_i - \phi_i < 0 & : \; \text{Group 2}
	\end{matrix}
\end{eqnarray}
where $\phi_i = \boldsymbol{\upphi}_i \cdot \mathbf{e}_x$. In other words, 
\begin{eqnarray}
    \begin{matrix}
	v_i \geq \phi_i & : \; \text{Group 1}\\
	v_i < \phi_i & : \; \text{Group 2}
	\end{matrix}
\end{eqnarray}

We now discuss an approach to estimate $\boldsymbol{\upphi}_i$. As mentioned before, we use information from the local neighbourhood to compute a \emph{neighbourhood parameter} $\widehat{\boldsymbol{\upphi}}_i$, which will serve as a surrogate for $\boldsymbol{\upphi}_i$. To define $\widehat{\boldsymbol{\upphi}}_i$, we make the following observations and assumptions: 
\begin{enumerate}[label=(\roman*)]
    \item an agent's motion is influenced predominantly by its immediate neighbours,
    \item a neighbour can either aid or oppose the intrinsic movement of the agent,
    \item the influence of a neighbour on an agent is dependent on the relative velocity of the neighbour, i.e. how fast the neighbour is approaching the position of the agent.
\end{enumerate}

Based on this, we define $\widehat{\boldsymbol{\upphi}}_i$ as follows:
\begin{eqnarray}
    \widehat{\boldsymbol{\upphi}}_i = \mu \left \langle (\mathbf v_j \cdot \mathbf e_{ji}) \, \mathbf e_{ji} \right \rangle _{j \in \mathcal{N}_i}  \label{eqn:phihat}
\end{eqnarray}

The parameter $ \widehat{\boldsymbol{\upphi}}_i$ is a vector capturing the net effect of neighbours (in the Voronoi neighbourhood $\mathcal{N}_i$) on the focal agent $i$. Agents interact as they approach each other; so the neighbour velocities are projected along the line joining the centres of $i$ and $j$. The projected velocity vectors are then averaged to get the net neighbourhood effect. $\mu$ is a scaling factor that ensures that $ \widehat{\boldsymbol{\upphi}}_i$ is of the same scale as $\mathbf v_i$. See Figure~\ref{fig:neigh-obs}(a) for an illustration.

One can estimate $\mu$ using a very simple scaling argument. Consider the case with the focal agent surrounded by 6 neighbours, in a regular hexagonal arrangement. In the extreme case, all of these neighbours are moving in the same direction, with the same velocity $\mathbf v$. In this case, $\left \langle (\mathbf v_j \cdot \mathbf e_{ji}) \, \mathbf e_{ji} \right \rangle _{j \in \mathcal{N}_i}$ will evaluate to $\frac{1}{6}(2 + 4 \cos \frac{\pi}{3}) \mathbf v = \frac{2}{3} \mathbf v$, giving $\mu = \frac{3}{2}$. Although a crude approximation, this value of $\mu$ works well in practice. See Discussion section for a comparison with an alternative, data-driven approach to estimating $\mu$.

It is important to note that the neighbourhood parameter is independent of the details of the simulation model. We use relative velocities of the neighbours as a proxy for inter-agent interactions; in other words, the neighbourhood parameter only depends on the overall phenomenology of the collective, and not on specific modelling assumptions.

\paragraph*{Classifying agents with the neighbourhood observer.}
In addition to computing the movement characteristics of just the agent $ v_i $, the neighbourhood observer calculates $ \phi_i $, which is a characteristic of its neighbourhood. As before, we use the time-window-averaged versions of these quantities.
    
Figure~\ref{fig:neigh-obs}(b) illustrates how the neighbourhood observer works. The plot shows the time-series of the (window-averaged) agent velocity $v_i^w$ for a Group~1 agent. When it encounters Group~2 agents in its path, it may get pushed, causing $v_i^w$ to dip below 0. The agent-only observer would then classify the agent---incorrectly---as Group~2: see blue shaded regions in the plot. However, the classification threshold for the neighbourhood observer is $\phi_i^w$ (blue dashed line in the plot), incorporates information about the neighbourhood of the agent, and can vary according to how much the agent is being pushed. Hence, the neighbourhood observer is able to correctly classify the agent as Group~1 even when $v_i^w <0$, i.e. the agent is moving in the negative x-direction.

\paragraph*{The neighbourhood observer improves performance on minority agents. } 
Accounting for neighbourhood effects significantly improves classification performance, especially for the minority group (Figure~\ref{fig:neigh-obs}(c), main panel). Even when only a single Group 2 agent is present, the neighbourhood observer can correctly classify the Group 2 agent ~60\% of the time. On the other hand, since a lone Group~2 agent is prone to getting pushed by Group~1 agents most of the time, simple observer almost always misclassifies it. There is a slight increase in the probability of misclassifying the majority group agents; but is small relative to the improvement achieved for the minority group.

Even in the high mobility scenario, where the performance of the simple observer is already high, the neighbourhood observer achieves significant improvements. In the low mobility scenario, where the agents are mostly jammed and there is very little relative movement, the improvement is relatively minor---see Section~\ref{sec:analysis} for a discussion about why this happens.

The neighbourhood observer can accurately estimate the level of heterogeneity across a wide range of parameters, even in cases where the simple observer under-estimates it. (Figure~\ref{fig:neigh-obs}(d), main panel). The heterogeneity estimate suffers when the mobility is very low, but the estimates are overall better than that of the simple observer. In particular, the neighbourhood observer is able to detect the presence of Group 2 agents at much smaller values of $N_r$ than the simple observer.

% \begin{itemize}
%     \item In general, classification performance improves significantly for the minority group.
%     \item The improvement is most pronounced for intermediate mobility levels, for asymmetric cases.
%     \item For low mobility; movement is severely restricted, so the improvement is not as significant. (see following sections for a discussion).
% \end{itemize}

% \paragraph*{Estimating heterogeneity with the neighbourhood observer}
% \begin{itemize}
%     \item Except when mobility is very low, the neighbourhood classifier is able to accurately estimate the level of heterogeneity.
%     \item For low mobility, the neighbourhood improves over the simple observer, but still under-estimates heterogeneity. However, it detects presence of heterogeneity much better than the simple observer.
% \end{itemize}

\subsection{\label{sec:analysis} Analysis of the observer algorithms}

\begin{figure}
    \centering
    \includegraphics[width=0.7\linewidth]{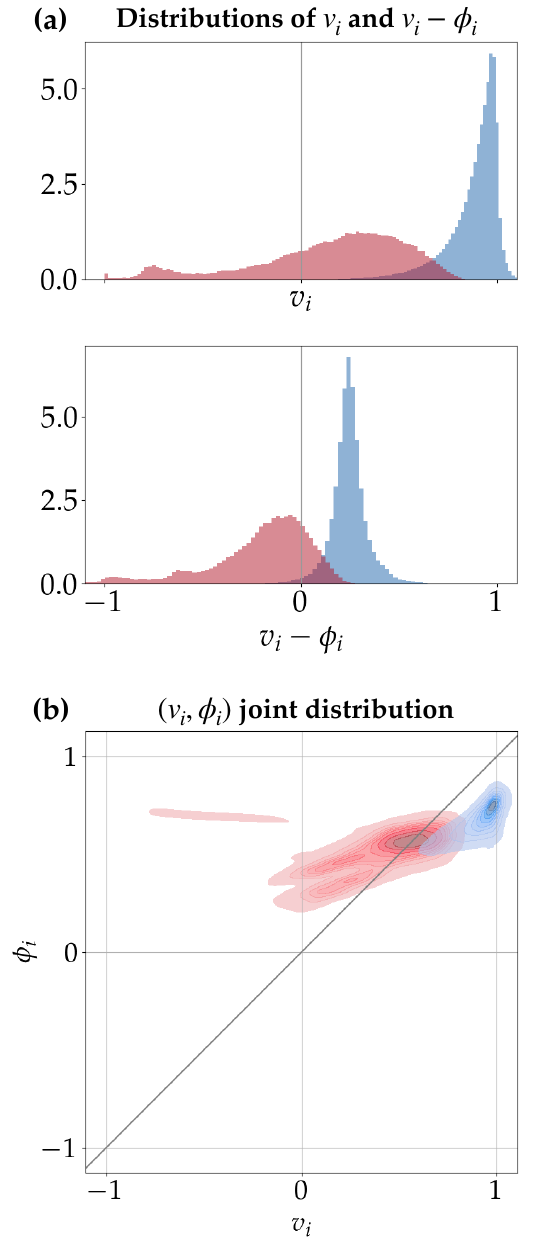}
    \caption{
    Illustration of how the neighbourhood classifier works.
    (a) Example distribution of individual horizontal velocities $v_i$ for Group 1 (blue) and Group 2 (red) agents. A large portion of the distribution for Group 2 (the minority group) lies on the positive side, leading to misclassifications.
    (b) Example distribution of $v_i - \phi_i$, the classification parameter used by the neighbourhood observer. The distributions are shifted so that the majority of the Group 2 distribution lies on the negative side.
    (c) The joint distribution of $v_i$ and $\phi_i$ for the two groups of agents. The classification boundary for the simple observer is the $v_i = 0$ vertical line, while the classification boundary for the neighbourhood observer is the $v_i = \phi_i$ diagonal line.
    }
    \label{fig:distributions}
\end{figure}

Figure~\ref{fig:distributions}(a) illustrates why the simple, velocity-based classification approach often fails. The top panel shows the histograms of the observed horizontal velocities ($v_i = \mathbf v_i \cdot \mathbf e_x$) of the two groups, for a typical scenario (asymmetric, intermediate mobility). The minority group  (Group 2) agents end up being pushed in the opposite direction frequently. As a result, the velocity distribution shifts to the right, with the bulk of the distribution lying on the positive sides. As a result, the simple observer, which classifies agents based on whether the observed velocity is positive or negative, misclassifies most of the minority group agents.

The bottom panel of Figure~\ref{fig:distributions}(a) illustrates how the neighbourhood observer gets around this problem. The distribution of $v_i - \phi_i$ for the minority group is more predominantly on the negative side, compared to the $v_i$ distribution. Therefore, the neighbourhood observer (which classifies agents based on $v_i \lessgtr \phi_i$ i.e. $v_i - \phi \lessgtr 0$) is able to classify the minority group agents much more accurately. Notice that the distribution of the majority group, although moves closer to the origin, stays predominantly on the positive side.

The observers can alternatively be viewed as a linear classifiers in the $(\phi_i, v_i)$ plane, as illustrated by Figure~\ref{fig:distributions}(b). The figure shows the joint distributions of $\phi_i$ and $v_i$ for the two different groups. The simple observer classifies agents based on $v_i$ and ignores $\phi_i$ altogether; so its decision boundary is the $v_i = 0$ line, i.e. a vertical line through the origin. Since the bulk of the minority group distribution lies to the right of this line, misclassifications will be high. 
On the other hand, the decision boundary for the neighbourhood observer is the $v_i = \phi_i$ line, i.e. a diagonal line with slope 1. The bulk of the minority group distribution is above this line, and will be classified correctly.

\begin{figure*}
    \centering
    \includegraphics[width=\textwidth]{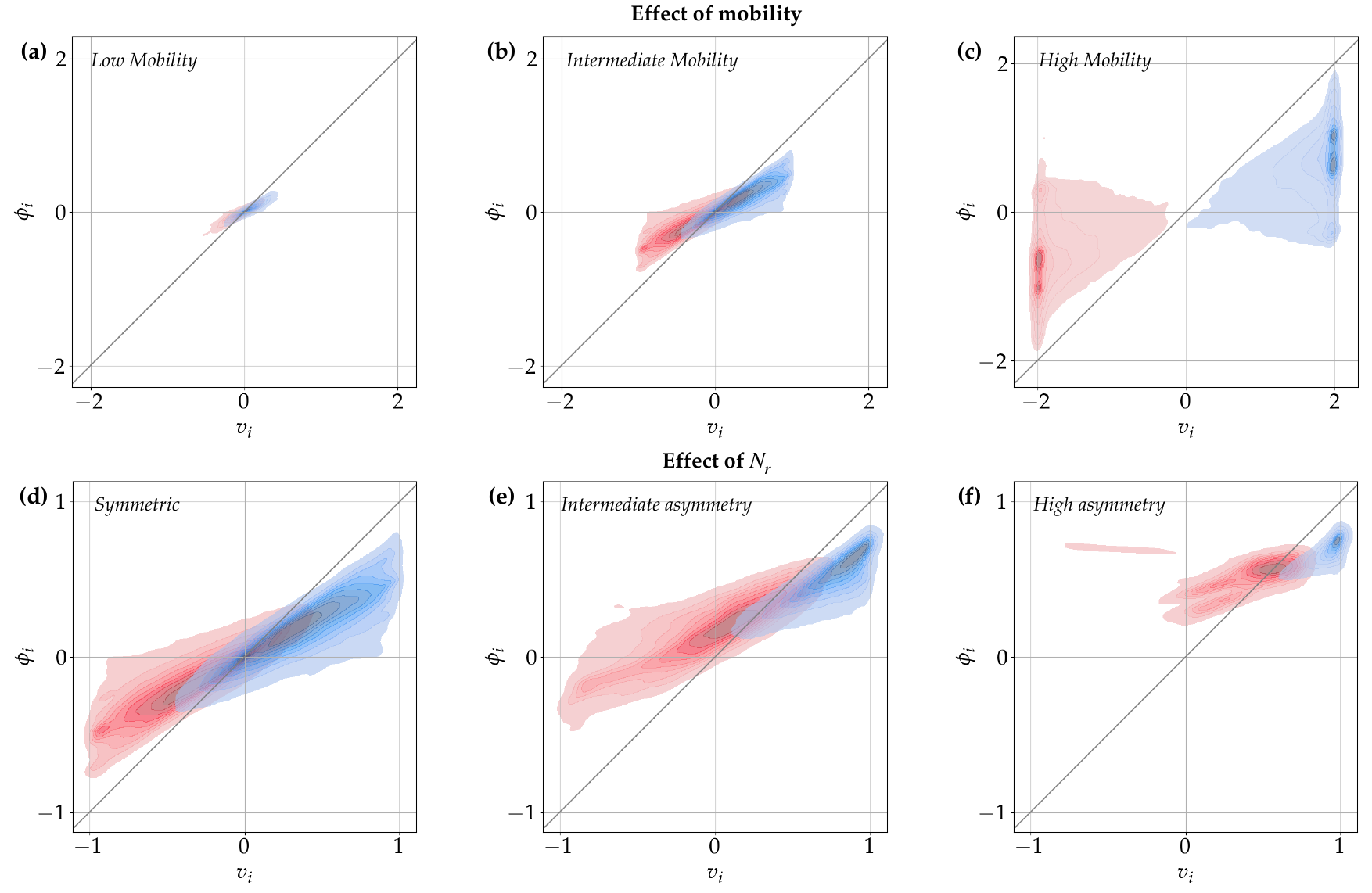}
    \caption{Effect of mobility and heterogeneity on the distributions and classification performance.
    (a-c) As mobility increases, the agents move more freely, causing the distributions to become well-separated, improving classification performance of both observers.
    (d-f) As asymmetry between the group sizes increases (i.e. $N_r$ decreases), the distributions shift towards the right, resulting in higher misclassifications for the smaller group by the simple observer. However, the shift is along the diagonal; which ensures that the performance of the neighbourhood observer does not drop significantly. }
    \label{fig:distribution-shifts}
\end{figure*}

\paragraph*{Effect of mobility and heterogeneity. }
The $(v_i, \phi_i)$ plane presents an elegant way to observe how mobility and heterogeneity affects the classification performance of the two observers, as illustrated in Figure~\ref{fig:distribution-shifts}.

The effect of mobility is quite clear: as mobility increases ($s_0$ increases and/or $\rho$ decreases), obstructions and jamming decreases and the agents move with higher velocities more often. This reflects as a separation of the Group 1 and Group 2 distributions on the $v_i, \phi_i$ plane (Figure~\ref{fig:distribution-shifts}(a-c)). At low mobility, there is a large amount of overlap between the two distributions, and both classifiers perform relatively poorly. On the other hand, at high mobility, the distributions are well separated and even the simple observer performs well.

The effect of heterogeneity on the distributions is perhaps more interesting. In the symmetric case ($N_r = 1/2$), there is no bias in the movement freedom for either group. This means that the distributions are arranged symmetrically on either sides of the $v_i = 0$ line, with their overlap determined by the level of mobility. As $N_r$ decreases and the collective becomes more asymmetric, the minority agents gets pushed more and the mixture velocity (i.e. the average velocity computed over all the agents) increases in the positive direction, causing both distributions to shift right. This degrades the performance of the simple observer.

However, when an agent is being pushed with some velocity, $\phi$, which is a proxy for the level of `push' on an agent by its neighbours, also increases proportionately: this means that in the $(v_i, \phi_i)$ plane, the shift happens diagonally, along the classification boundary of the neighbourhood observer: the regions of the distributions that are above and below the classification boundary stay roughly the same. This is the reason why asymmetry does not degrade the performance of the neighbourhood observer too much.

\section{Discussion}

\paragraph*{Summary. }
In this article, we set out to understand how well can we infer individual properties of agents in a heterogeneous collective by observing their movement. We used a bidisperse collective as a context and studied how the classification process of an observer is connected to the physics of collective motion. We began with a simple observation/classification technique based on the measured velocity of of the individual agents, independent of their neighbourhood. We found that the simple observer underestimates the level of heterogeneity of the collective, and systematically misclassifies the minority group agents more often than the majority group.
	
To improve classification, we developed an observer that classifies an agent based on information not just about itself, but also about its neighbours. Although simple in principle, this neighbourhood observer does something quite non-trivial: it distinguishes between a Group~1 and a Group~2 agent even when they are moving in the same direction with identical velocities. The fact that the neighbourhood observer takes into account the influence of the neighbouring agents for classification helps it to `read the mind' of the focal agent by distinguishing when it is moving on its own volition versus when it is being pushed. This resulted in better estimates of the level of heterogeneity, and improved classification for the minority group.

\paragraph*{Data-driven approaches to the classification problem.} The $(v_i, \phi_i)$ classifier that the neighbourhood observer employs is not data-driven: the classification boundary (which is decided by $\mu$) is not obtained by a data fit, but is instead derived from scale considerations. This makes our approach applicable in scenarios where labelled data is not easy to obtain. This also makes our classifier readily interpretable, as it is inspired by the physics of the problem. The assumptions the classification algorithm makes about the underlying governing dynamics are minimal. Specifically, the neighbourhood parameter is computed based on neighbour velocities only, and is independent of specific details of how the agents interact.
	
If labelled data (i.e. data where the group memberships of agents are known) is readily available, we can learn $\mu$ in a data-driven manner, by fitting a linear classifier. When a classifier is fit in this manner, it will not be susceptible to unfulfilled assumptions in computing $\mu$. However, the data-driven classifier offers no improvements over the simple linear classifier (see Appendix~\ref{sec:svm-obs}), underscoring the accuracy of the scaling argument.
	
An orthogonal approach is to build a purely data-driven classifier, for example, based on neural networks. One could build a neural-net classifier that takes as input the positions and velocities of the focal agent and its neighbours, and makes predictions. Given enough labelled data, such classifiers can learn high-quality, low-dimensional feature representations that can make effective predictions. To be effective, such neural network classifiers should be informed and constrained by the physics and inherent symmetries of the problem. In our case, it means that the neural network should respect the rotational and translational invariances of the system, and should be invariant to any permutation of the agent ordering. Constraints such as these should be built in to the neural network design. There is an emerging body of research in graph neural networks \cite{zhou2020graph, bronstein2021geometric} and physics-informed deep learning \cite{karniadakis2021physics, raissi2019physics} which can be explored in this regard. Building neural network classifiers for collective dynamics and studying their feature representations is an exciting research direction.

\paragraph*{Towards a general inference framework for heterogeneous collectives.}

There are several ways in which the classification algorithm can be improved further. Firstly, notice that the ultimate goal would be to recover $\mathbf v_{0, i}$, for example by finding $\boldsymbol \upphi_i$ such that $\mathbf v_i - \boldsymbol \upphi_i = \mathbf v_{0, i}$. Our estimated neighbourhood parameter doesn't quite achieve this: however, it gets the sign correctly in general, i.e $\mathrm{sgn} (v_i - \phi_i) = \mathrm{sgn}(v_{0, i})$, which is sufficient for classification.

Recall that the neighbourhood parameter for an agent $i$ was computed based on the mean of the velocities of the Voronoi neighbours of $i$. Potential ways to improve this would be to use a different neighbourhood, use a weighted average weighted by the distance to the neighbour, or use more complicated functions of the velocity.

One could also use information about the spatial arrangement of the neighbours of $i$. For example, it is conceivable that the fore-aft asymmetry pattern in the neighbourhood of $i$ will be different when $i$ is being pushed versus when $i$ is moving freely. Quantifying these spatial patterns could be an effective approach to recover intrinsic motion.

Finally, note that both the observers treat the time windows as independent: when making predictions for a given time window, the observers do not use information from previously observed windows. One could conceive an observer that uses historical information and updates beliefs with time, perhaps in a Bayesian fashion. We do not explore this approach because our goal is to explore techniques that work in a nearly \emph{real-time} manner with minimal observations, and to understand how instantaneous and short-term dynamics in the system affects our ability to classify agents. Another way to incorporate temporal information would be to look for patterns in the higher derivatives (e.g. acceleration) of motion of both the focal agent and the neighbourhood.

\paragraph*{Conclusion}
% Classifying agents in a real collective is a much harder problem to study in comparison to the model system we investigate in this article. We assume the composite collective to be \textit{bidisperse}---consisting of two groups of agents and \textit{uniform}---indistinguishable from each other. However, agents in a real collective are far more heterogeneous, where there may not be well-defined, distinct groups like we assume in this study. We also assume that observer knows the collective to be bidisperse. In fact, when the observer attempts to infer the \textit{rule} that the agents follow, it never uncovers the same value for $v_{0,i}$. Almost always, the observer infers a different value for each agent in a certain group. Simply put, even a uniform group of agents are identified as non-uniform due to the differences in the conditions they experience as they move in the collective \cite{Schumacher2017Heterogeneity}.
	
It is worth reiterating that the goal of this study is to understand the ways in which the microscopic properties of a heterogeneous collective affect the classification performance of an observer that is only privy to observables such as agent positions and velocities. We study this in the context of a model system, which is an idealised version of a heterogeneous collective in the real world. For this reason, hunting for the best data-driven classifier to solve this specific problem may not be very useful; such a classifier will not be transferable to a real-world collective as real-world collectives can be poly-disperse and can involve many complex interactions not included in our model. Hence, our main goal was to understand how the collective dynamics can affect classification. A clear understanding of this relationship is essential to solve similar problems in real-world collectives. Our phenomenological approach to the classification problem is a first step towards hybrid techniques, where a data-driven approach is combined with domain-specific understanding of the collective dynamics to build better observer models.

\section*{Acknowledgements}
The authors thank Jason Ryan Picardo, Ganapathy Ayappa, Vishwesha Guttal, Vivek Jadhav, Basil Thurakkal and Harishankar Muppirala for critical readings of the manuscript. The authors thank the DST INSPIRE faculty award for the funding (grant no. DST/INSPIRE/04/2017/002985) \\

\section*{Code availability}
The simulation and analysis code used in this work is available at \url{https://github.com/arshednabeel/ObservingAcollective}

\appendix

\section{\label{sec:svm-obs} Appendix: Classifying agents with a data-driven classifier}

\begin{figure*}
    \centering
    \includegraphics[width=\textwidth]{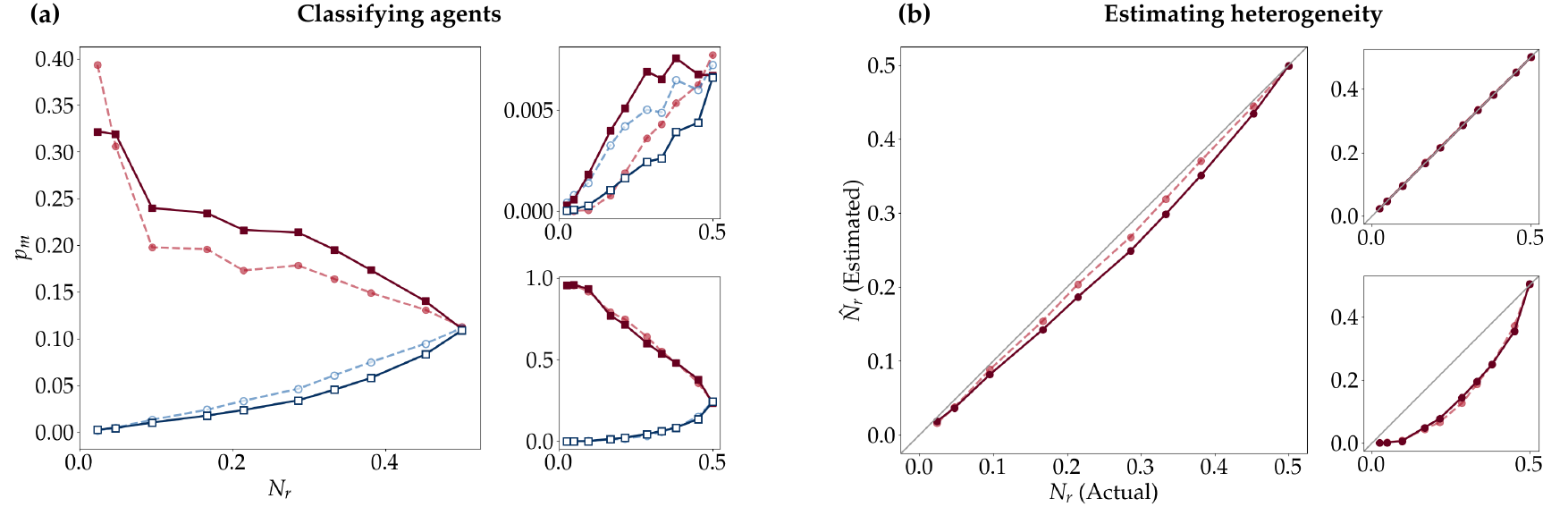}
    \caption{Comparing the (physics-informed) neighbourhood classifier to a data-driven classifier: the physics-informed neighbourhood observer performs as good as a data-driven classifier fine-tuned to the dataset.
    (a) The probability of misclassifying agents $p_m$ for the minority (red, filled squares) and majority (blue, open squares) groups, as a function of the number ratio $N_r$, for different levels of agent mobility (main: intermediate mobility, $\rho = 0.46, s_0 = 1$, top right: high mobility, $\rho = 0.31, s_0 = 2$, bottom right: low mobility, $\rho = 0.58, s_0 = 0.75$).
    The performance curves of the physics-informed neighbourhood observer are shown as dotted lines. There is no significant improvement with the data-driven classifier.
    (b) The estimated value of $N_r$, which is a measure of estimated asymmetry, as a function of true $N_r$, for different levels of agent mobility. The performance curves of the (analytically derived) neighbour observer are shown as dotted lines. 
    }
    \label{fig:svm-obs}
\end{figure*}

In Section~\ref{sec:neigh-obs}, we used a scaling argument to analytically estimate $\mu$. Alternatively, if labelled data is available, $\mu$ can be estimated in a data-driven manner by fitting a linear classifier.

Section~\ref{sec:neigh-obs} defined the neighbourhood classifier with a classification criterion $v_i \lessgtr \phi_i$. Let $\phi_i = \mu \varphi_i$ where $\varphi_i$ corresponds to the unscaled mean from Equation~\ref{eqn:phihat}. In this formulation, the classification boundary (see Section~\ref{sec:analysis} and Figure~\ref{fig:distributions}) corresponds to a line of slope $1/\mu$ in the $(v_i, \varphi_i)$ plane. With labelled data, one could fit a linear classifier (e.g. a support vector machine) to the data to obtain the `optimal' $\mu$ for the data-set.

We trained an SVM to distinguish between Group~1 and Group~2 agents in this way. Since we are only interested in learning a scaling factor $\mu$, we enforce that the intercept of the classifier be zero. A separate model was fit for each set of parameters ($N_r$, $s_0$ and $\rho$), with data pooled over all realizations and time-points. Since to our goal is to find the theoretically optimal $\mu$, we did not use separate training and test sets, performance reported is from the same dataset used for training: this is an upper bound for reported accuracy numbers with separate training and test sets.

Figure~\ref{fig:svm-obs} compares the performance of the analytically derived (as in Section~\ref{sec:neigh-obs}) and data-driven classifiers. The data-driven approach does not show any significant improvement in classification performance---in terms of classifying agents as well as estimating heterogeneity, the SVM-based classifier performs nearly identically as our analytically derived classifier. This suggests that our estimate of $\mu$ is close to optimal.

\bibliography{references.bib,library2.bib}% Produces the bibliography via BibTeX.

\end{document}